\documentclass[wcp]{jmlr}

\usepackage{longtable}

\usepackage{booktabs}

\usepackage{graphicx}
\usepackage{multirow, bigdelim}
\usepackage{booktabs}

\usepackage{tikz}
\usepackage{afterpage}
\usetikzlibrary{shapes.geometric, arrows, calc}

\tikzstyle{data} = [rectangle, rounded corners, minimum width=3cm, text width=3cm, minimum height=1cm, text centered, draw=black, fill=black!0]

\tikzstyle{model} = [rectangle, rounded corners, minimum width=3cm, text width=3cm, minimum height=1.5cm, text centered, draw=black, fill=black!20]

\tikzstyle{arrow} = [thick, ->, >=stealth]

\jmlrvolume{77}
\jmlryear{2017}
\jmlrworkshop{ACML 2017}

\title[Probability Calibration Trees]{Probability Calibration Trees}

\author{\Name{Tim Leathart} \Email{tml15@students.waikato.ac.nz}\\
   \Name{Eibe Frank} \Email{eibe@cs.waikato.ac.nz}\\
   \Name{Geoffrey Holmes} \Email{geoff@cs.waikato.ac.nz}\\
   \addr Department of Computer Science, University of Waikato
   \AND
   \Name{Bernhard Pfahringer} \Email{b.pfahringer@auckland.ac.nz}\\
   \addr Department of Computer Science, University of Auckland
}

\editors{Yung-Kyun Noh and Min-Ling Zhang}

\begin{document}

\maketitle

\begin{abstract}
Obtaining accurate and well calibrated probability estimates from classifiers is useful in many applications, for example, when minimising the expected cost of classifications. Existing methods of calibrating probability estimates are applied globally, ignoring the potential for improvements by applying a more fine-grained model. We propose probability calibration trees, a modification of logistic model trees that identifies regions of the input space in which different probability calibration models are learned to improve performance. We compare probability calibration trees to two widely used calibration methods---isotonic regression and Platt scaling---and show that our method results in lower root mean squared error on average than both methods, for estimates produced by a variety of base learners.

\end{abstract}
\begin{keywords}
Probability calibration, logistic model trees, logistic regression, LogitBoost
\end{keywords}

\section{Introduction}
In supervised classification, assuming uniform misclassification costs, it is sufficient to predict the most likely class for a given test instance. However, in some applications, it is important to produce an accurate probability distribution over the classes for each example. While most classifiers can produce a probability distribution for a given test instance, these probabilities are often not well \textit{calibrated},~\textit{i.e.,} they may not be representative of the true probability of the instance belonging to a particular class. For example, for those test instances $x$ that are assigned a probability of belonging to class $j$, $P(y=j \mid x) = 0.8$, we should expect approximately $80\%$ to actually belong to class $j$. 

The class probability estimates produced by a classifier can be adjusted to more accurately represent their underlying probability distributions through a process called probability calibration. This is a useful technique for many applications, and is widely used in practice. For example, in a cost-sensitive classification setting, accurate probability estimates for each class are necessary to minimise the total cost. This is because the decision is made based on the lowest expected cost of the classification, $\sum_{i=1}^m C(y'=i\mid y=j)P(y=j\mid x)$, where $m$ is the number of classes and $C(y'=i\mid y=j)$ is the cost of classifying an instance as class $i$ when it belongs to class $j$, rather than simply the most likely class. It can also be important to have well calibrated class probability estimates if these estimates are used in conjunction with other data as input to another model.  Lastly, when data is highly unbalanced by class, probability estimates can be skewed towards the majority class, leading to poor scores for metrics such as $F_1$.

The most prevalent methods for probability calibration are Platt scaling~\citep{platt1999probabilistic} and isotonic regression~\citep{zadrozny2001obtaining}. These methods work well, but they assume that probability estimates should be calibrated in the same fashion in all regions of the input space. We hypothesise that in some cases, this assumption leads to poor probability calibration and that a more fine-grained calibration model can yield superior calibration overall. In this work we propose probability calibration trees, a novel probability calibration method based on logistic model trees~\citep{landwehr2005logistic}. Probability calibration trees identify and split regions of the instance space in which different probability calibration models are learned. We show that these localised calibration models often produce better calibrations than a single global calibration model.

This paper is structured as follows. In Section~\ref{sec:probability_calibration_methods} we give an overview of the two existing probability calibration methods mentioned above. Section~\ref{sec:logistic_model_trees} briefly introduces logistic model trees. Then, in Section~\ref{sec:probability_calibration_trees}, we explain our method of inducing probability calibration trees, and discuss how inference is performed. In Section~\ref{sec:experiments}, we present experiments that we performed to test the effectiveness of our proposed technique. Finally, we conclude and discuss future work in Section~\ref{sec:conclusion}.

\section{\label{sec:probability_calibration_methods}Probability Calibration Methods}
Probability calibration is widely applied in practice. In this section, we discuss Platt scaling and isotonic regression, the most commonly used methods for probability calibration. We also briefly describe some other, more recent approaches to probability calibration.

\subsection{Platt Scaling}
\cite{platt1999probabilistic} introduce a method of probability calibration for support vector machines (SVMs) called Platt scaling. In this method, predictions in the range $[-\infty, +\infty]$ are passed through a sigmoid function to produce probability estimates in the range $[0, 1]$. The sigmoid function is fitted with logistic regression. Platt scaling is only directly applicable to a two class problem, but standard multiclass classification techniques such as the one-vs-rest method~\citep{rifkin2004defense} can be used to overcome this limitation. The logistic regression model must be trained on an independent calibration dataset to reduce overfitting. Before the logistic regression model is fitted, Platt suggests a new labeling scheme where instead of using $y_+ = 1$ and $y_- = 0$ for positive and negative classes, the following values are used:

	\begin{equation}
		y_+ = \frac{N_+ + 1}{N_+ + 2},\quad y_- = \frac{1}{N_- + 2},
	\end{equation}
	
	where $N_+$ and $N_-$ are the number of positive and negative examples respectively. This transformation follows from applying Bayes' rule to a model of out-of-sample data that has a uniform prior over the labels~\citep{platt1999probabilistic}. 
	
	Although Platt scaling was originally proposed to scale the outputs of SVMs, it has been shown to work well for boosted models and naive Bayes classifiers as well~\citep{niculescu2005predicting}.

\subsection{Isotonic Regression}
\cite{zadrozny2001obtaining} use a method based on isotonic regression for probability calibration for a range of classification models. Isotonic regression is more general than Platt scaling because no assumptions are made about the form of the mapping function, other than it needs to be monotonically increasing (isotonic). A non-parametric piecewise constant function is used to approximate the function that maps from the predicted probabilities to the desired values. The mapping function with the lowest mean squared error on the calibration data can be found in linear time using the pair-adjacent violators algorithm~\citep{ayer1955empirical}.

Like Platt scaling, an independent calibration set is used to fit the isotonic regression mapping function to avoid unwanted bias. Isotonic regression can only be used on a two-class problem, so multiclass classification techniques must be used when applying it in a multiclass setting.

\subsection{\label{sec:other_probability_calibration_methods}Other approaches}
\cite{ruping2006robust} show that both Platt scaling and isotonic regression are greatly affected by outliers in the probability space. In their research, Platt scaling is modified using methods from robust statistics to make the calibration less sensitive to outliers. \cite{jiang2011smooth} propose to construct a smooth, monotonically increasing spline that interpolates between a series of representative points chosen from a isotonic regression function. \cite{zhong2013accurate} incorporate manifold regularisation into isotonic regression to make the function smooth, and adapt the technique to be better suited to calibrating the probabilities produced by an ensemble of classifiers, rather than a single classifier. 

\section{\label{sec:logistic_model_trees}Logistic Model Trees}
Our probability calibration method is derived from the algorithm for learning logistic model trees~\citep{landwehr2005logistic}. Logistic model trees, on average, outperform both decision trees and logistic regression. They also perform competitively with ensembles of boosted decision trees while providing a more interpretable model. Simply put, logistic model trees are decision trees with logistic regression models at the leaf nodes, providing an adaptive model that can easily and automatically adjust its complexity depending on the training dataset. For small, simple datasets where a linear model gives the best performance, this is simply a logistic regression model (\textit{i.e.}, a logistic model tree with only a single node). For more complicated datasets, a more complex tree structure can be built.

While a logistic model tree is grown, each split node is considered a candidate leaf node, so a logistic model is associated with every node in the tree. Instead of fitting a logistic regression model from scratch at each node, the LogitBoost algorithm~\citep{friedman2000additive}, applying simple linear regression based on a single attribute as the weak learner, is used to incrementally refine logistic models that have already been learned at previous levels of the tree. Cross-validation is used to determine an appropriate number of boosting iterations. This results in an additive logistic regression model of the form

\begin{equation}	\label{eqn:logistic}
	P(y=j\mid x) = \frac{e^{F_j(x)}}{\sum_{i=1}^{m} e^{F_i(x)}} \textrm{\quad where \quad} \sum_{i=1}^{m} F_i(x) = 0.
\end{equation}

\noindent Here, $m$ is the number of classes, $F_i(x) = \sum_{k=1}^l f_{ik}(x)$, $l$ is the number of boosting iterations, and each $f_{ik}$ is a simple linear regression function. 

The C4.5 algorithm is used to construct the basic tree structure before logistic models are fit to the nodes. After the tree has been grown, it is pruned using cost-complexity pruning~\citep{breiman1984classification}, which considers both the training error and the complexity of the tree. Missing values are replaced with the mean (for numeric attributes) or mode (for categorical attributes). Categorical attributes are converted to binary indicator variables for the logistic models.

\section{\label{sec:probability_calibration_trees}Probability Calibration Trees}

\begin{figure}[t]
	\begin{tikzpicture}[node distance=2cm]
	
		\node (base) [model] {Base Model};
		
		\node (original_data2) [data, right of=base, xshift=-2cm, yshift=2cm] {Original Data};
		\node (estimates) [data, right of=base, xshift=2cm] {Output Scores};
		
		\node (pct) [model, right of=base, xshift=6cm] {Probability Calibration Tree};
		
		\node (calibrated) [data, right of=pct, xshift=2cm] {Calibrated Probabilities};
		
		\draw[arrow] (original_data2) -- (base);
		\draw[arrow] (base) -- (estimates);
		\draw[arrow] (estimates) -- (pct);
		\draw[arrow] (original_data2) -| (pct);
		\draw[arrow] (pct) -- (calibrated);
		
	\end{tikzpicture}
	
	\caption{\label{fig:process} The process of obtaining calibrated probabilities from a probability calibration tree.}
\end{figure}
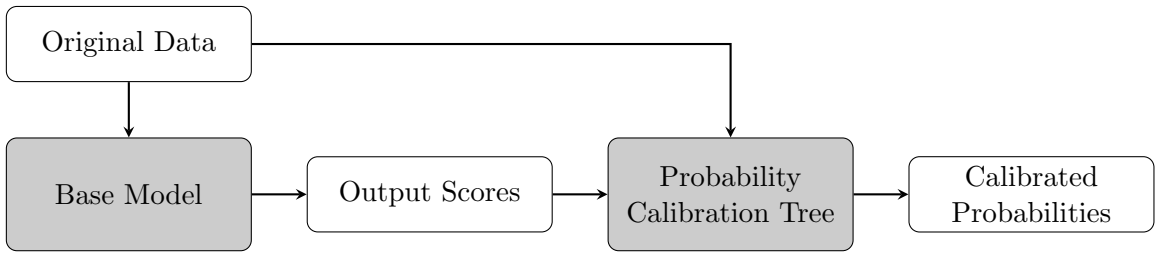

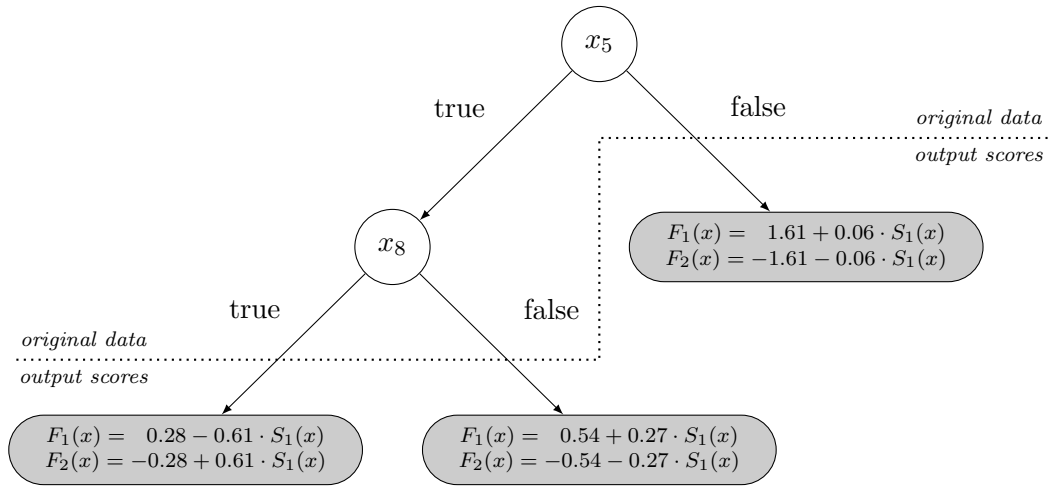
\begin{figure}[t]
	\centering
		\begin{tikzpicture}[
			edge from parent/.style={draw=black,-latex},
		]
		\tikzstyle{level 1}=[sibling distance = 55mm, level distance=27mm]
		\tikzstyle{level 2}=[sibling distance = 55mm, level distance=27mm]
		\usetikzlibrary{shapes}
		
		\node[circle,draw,minimum size=1cm](root){$x_5$}
		  child
		  {
		  	node[circle,draw,minimum size=1cm](left){$x_8$}
		  	child
		  	{
		  		node[rounded rectangle,draw,fill=black!20](leftleft)
		  		{
		  			\scriptsize
		  			\begin{tabular}{c} 
						$F_1(x) = ~~0.28 - 0.61 \cdot S_1(x)$ \\
						$F_2(x) = -0.28 + 0.61 \cdot S_1(x)$ 
		  			\end{tabular}
				}
		  	}
		  	child
		  	{
		  		node[rounded rectangle,draw,fill=black!20](leftright)
		  		{
		  			\scriptsize
			  		\begin{tabular}{c} 
						$F_1(x) = ~~0.54 + 0.27 \cdot S_1(x)$ \\
						$F_2(x) = -0.54 - 0.27 \cdot S_1(x)$ 		
		  			\end{tabular}
		  		}
		  	}
		  }
		  child
		  {
	    		node[rounded rectangle,draw,fill=black!20](right)
	    		{
		  			\scriptsize
		    		\begin{tabular}{c} 
						$F_1(x) = ~~1.61 + 0.06 \cdot S_1(x)$ \\
						$F_2(x) = -1.61 - 0.06 \cdot S_1(x)$
		  			\end{tabular}
	    		}
	    	};
	      
	      \node[align=left] at (-6.8,-3.95) { \scriptsize \textit{original data} };
	      \node[align=left] at (-6.8,-4.45) { \scriptsize \textit{output scores} };
	      
	      \node[align=left] at (5.1,-1) { \scriptsize \textit{original data} };
	      \node[align=left] at (5.1,-1.5) { \scriptsize \textit{output scores} };
	      
	      \draw[thick,dotted] (-7.75, -4.2) -- (0, -4.2) -- (0, -1.25) -- (6, -1.25);
	      
	      \begin{scope}[nodes = {draw = none}]
		    \path (root) -- (left) node [near start, left]  {true \quad\quad};
		    \path (root) -- (right) node [near start, right]  {\quad\quad false};

		    \path (left) -- (leftleft) node [near start, left]  {true \quad\quad};
		    \path (left) -- (leftright) node [near start, right]  {\quad\quad false};		    
		  \end{scope}
	\end{tikzpicture}
	
	\caption{\label{fig:pct}A probability calibration tree for the outputs of an SVM with an RBF kernel ($C = 10, \gamma = 0.01$) on the \texttt{RDG1} dataset. \texttt{RDG1} is a small two-class dataset with 10 binary attributes, and can be generated in the WEKA software using the eponymous data generator. $x_5$ and $x_8$ are attributes in the original data, while $S_1(x)$ is the output score of the SVM. The functions $F_i(x)$ compute the calibrated log-odds estimate of $x$ belonging to class $i$, and must sum to zero. The final calibrated probabilities are computed with Equation~\ref{eqn:logistic}.}	
\end{figure}

Probability calibration trees are built using a similar algorithm to logistic model trees except they make use of two input datasets---(a) the original training data, and (b) the associated output scores from the base classifier that we want to calibrate such as probability estimates or SVM outputs~(Figure~\ref{fig:process}). The original training data---part (a) of the input data---is used to build the basic tree structure using the C4.5 algorithm, and the output scores---part (b) of the input data---are used to train the logistic models using LogitBoost. In this manner, a probability calibration tree performs Platt scaling in different regions of the input space when it is advantageous to do so, but uses a global Platt scaling model if this gives better performance. Therefore, we expect probability calibration trees to outperform or equal the performance of global Platt scaling. 
An example of a probability calibration tree is shown in Figure~\ref{fig:pct}. 

\subsection{Training Probability Calibration Trees}
At a high level, the process of training a probability calibration tree is as follows:

\begin{enumerate}
	\item Grow a decision tree from the original attributes, creating leaf nodes when some stopping criterion is met.
	\item For each node, train logistic regression models on the output scores of the training instances at that node.
	\item Prune the tree to minimise error.
\end{enumerate}

As in logistic model trees, the LogitBoost algorithm in conjunction with simple linear regression is used to train the logistic models. Each node uses the logistic model in its parent node as a `warm start' for the boosting process, but only the subset of instances present in the child node are used for future boosting iterations. We use the same stopping criteria for the growing process as logistic model trees, which is to create leaf nodes when fewer than 15 training instances are present at the node.

Similarly to logistic model trees, pruning is an important step in the fitting process. Logistic model trees are pruned to minimise the number of classification errors. However, probability calibration trees are intended to produce good probability estimates rather than classification accuracy. Therefore, we prune subtrees from the model until the root mean squared error (RMSE) of the calibrated probability estimates cannot be reduced further, as this is a better proxy for the quality of probability estimates than $0$-$1$ loss. The RMSE is the square root of the Brier score~\citep{brier1950verification} divided by the number of classes:

\begin{equation}
	\textrm{RMSE} = \sqrt{\frac{1}{nm}\sum_{i=1}^n \sum_{j=1}^m (p_{ij} - y_{ij})^2 }
\end{equation}

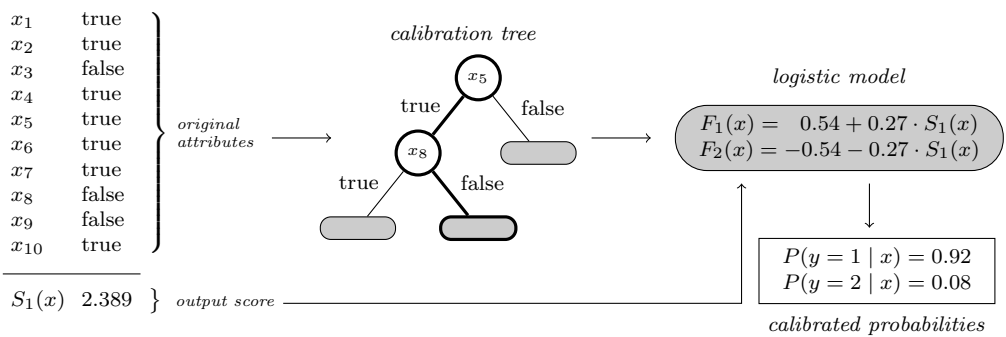
\begin{figure}[t!]
	\scriptsize
	\begin{minipage}[t]{0.9\linewidth}
		\begin{tikzpicture}[
				baseline,
				norm/.style={style={black,thin,draw}, edge from parent/.style={black,thin,draw}},
			    emphpath/.style={edge from parent/.style={black,very thick,draw}},
			    emph/.style={style={black,very thick,draw}}
			]

			\tikzstyle{level 1}=[sibling distance = 16mm, level distance=10mm]
			\tikzstyle{level 2}=[sibling distance = 16mm, level distance=10mm]
			\usetikzlibrary{shapes}

			\node at (0,0) 
			{
				\bgroup
					\setlength\tabcolsep{3pt}
					\begin{tabular}{lll}
						$x_1$ & true & \rdelim\}{10}{3mm}[\tiny 
						\begin{tabular}{l}
							\textit{original} \\
							\textit{attributes}
						\end{tabular}] \\
						$x_2$ & true & \\
						$x_3$ & false &\\
						$x_4$ & true &\\
						$x_5$ & true &\\
						$x_6$ & true &\\
						$x_7$ & true &\\
						$x_8$ & false &\\
						$x_9$ & false &\\
						$x_{10}$ & true &\\				
						\noalign{\vskip 2mm} 
						\cline{1-2}
						\noalign{\vskip 2mm}    
						$S_1(x)$ & 2.389 & \rdelim\}{1}{3mm}[\tiny \textit{~~output score}] \\ 
						&& \\ \\ \\
					\end{tabular}
				\egroup
			};
		
			\node (a1) at (3.35, 0.8) {};
			\node (b1) at (2.35, 0.8) {};
			\draw[->, thin] (b1) edge (a1);
			
			\node[emph,circle,draw,minimum size=0.4cm](root) at (5.2,1.6) {\tiny $x_5$}
				child[emph,emphpath]
				{
					node[circle,draw,minimum size=0.4cm](left){\tiny $x_8$}
					child[norm]
					{
						node[norm, rectangle,rounded corners,draw,fill=black!20,minimum height=0.3cm,minimum width=1cm](leftleft){}
					}
					child[emph,emphpath]
					{
						node[rectangle,rounded corners,draw,fill=black!20,minimum height=0.3cm,minimum width=1cm](leftright){}
					}
				}
				child[norm]
				{
					node[rectangle,rounded corners,draw,fill=black!20,minimum height=0.3cm,minimum width=1cm](right){}
				};

	      \begin{scope}[nodes = {draw = none}]
		    \path (root) -- (left) node [near start, left]  {true \quad};
		    \path (root) -- (right) node [near start, right]  {\enskip false};

		    \path (left) -- (leftleft) node [near start, left]  {true \quad};
		    \path (left) -- (leftright) node [near start, right]  {\enskip false};		    
		  \end{scope}

			\node at (5, 2.2) { \textit{\scriptsize calibration tree} };
	
			\node (A) at (7.6, 0.8) {};
			\node (B) at (6.6, 0.8) {};
			\draw[->, thin] (B) edge (A);
		
			\node[rectangle,rounded corners=12pt,draw,fill=black!20] at (10, 0.8)
			{
				\begin{tabular}{c} 					
					$F_1(x) = ~~0.54 + 0.27 \cdot S_1(x)$ \\ 		
					$F_2(x) = -0.54 - 0.27 \cdot S_1(x)$ 
	  			\end{tabular}
			};
			
			\node (a3) at (2.5, -1.4) {};
			\node (c3) at (8.7, 0.3) {};
			\draw[->, thin] (a3) -| (c3);

			\node (a4) at (10.4, -0.5) {};
			\node (b4) at (10.4, 0.3) {};
			\draw[->, thin] (b4) -- (a4);

			\node at (10, 1.6) { \textit{\scriptsize logistic model} };

			\node[draw, rectangle] at (10.5, -0.97)
			{
				\begin{tabular}{c}
					$ P(y = 1 \mid x) = 0.92 $ \\
					$ P(y = 2 \mid x) = 0.08 $ 
				\end{tabular}
			};
			
			\node at (10.5, -1.7) { \textit{\scriptsize calibrated probabilities} };

		\end{tikzpicture}
	\end{minipage}

	\caption{\label{fig:inference}The process of gaining calibrated probabilities for an instance $x$ from the calibration tree from Figure~\ref{fig:pct}. $x_1, x_2, \dots, x_{10}$ are the original attributes of $x$, and $S_1(x)$ is the output score from the SVM. First, the original attributes are used to select a logistic model from a leaf node in the calibration tree. Then, the output score $S_1(x)$ is used in the logistic model to produce calibrated probability estimates.}
\end{figure}

where $n$ is the number of instances, $m$ is the number of classes, $p_{ij}$ is the predicted probability that instance $i$ is of class $j$, and $y_{ij}$ is $1$ if instance $i$ actually belongs to class $j$, and $0$ otherwise. The CART pruning strategy based on cost-complexity is applied, which uses cross-validation to estimate error. Likewise, in probability calibration trees, the number of boosting iterations to use for LogitBoost is chosen via a cross-validated hyperparameter search optimising for RMSE, unlike in logistic model trees where this hyperparameter is optimised based on classification accuracy. The number of boosting iterations is determined once at the root node of the tree. This number is then applied at each node.

Logistic regression assumes a linear relationship between its input and the log-odds of the class probabilities which are output. When the output scores used as input to the probability calibration tree are probability estimates rather than SVM scores, we can decrease the error of the logistic models at the leaf nodes of the probability calibration tree by first transforming each of the input class probabilities $p_j$ into their log-odds $z_j$ before passing them to the probability calibration tree:

	\begin{equation} \label{eqn:logodds}
		z_j = \ln \bigg( \frac{p_j}{1-p_j} \bigg)
	\end{equation}
	
\noindent This assumes that there is a linear relationship between the log-odds of the original probability estimates and the log-odds of the calibrated probability estimates, because logistic regression models the \textit{log-odds} of the class probabilities---not the probability estimates themselves---as a linear combination of the input variables.

LogitBoost can build logistic regression models for multi-class problems, so a useful feature of probability calibration trees is that they are directly applicable to multiclass problems; there is no need to use a multiclass technique like one-vs-rest.

As with other probability calibration methods, it is important to train probability calibration trees on a held-out validation set to avoid overfitting. In all of our experiments, we obtain probability estimates with internal cross validation. Internal 5-fold cross-validation is used to collect class probability estimates from the base learner and corresponding true class labels for each held out instance.

\subsection{Inference in Probability Calibration Trees}

The process of calibrating probabilities for test instances with probability calibration trees is depicted in Figure~\ref{fig:inference}. When using the tree to compute calibrated probabilities for test instances, the instances are passed down the tree based on their original attributes. When the test instance reaches a leaf node, the probability estimates for that instance, obtained from the base model, are calibrated with the logistic regression model at that leaf node.

\subsection{\label{sec:artificial}An Artificial Example}

\begin{figure}[t]
	\centering
		{
		\setlength{\fboxsep}{0pt}%
		\setlength{\fboxrule}{1pt}%
        \fbox{\includegraphics[width=0.25\textwidth]{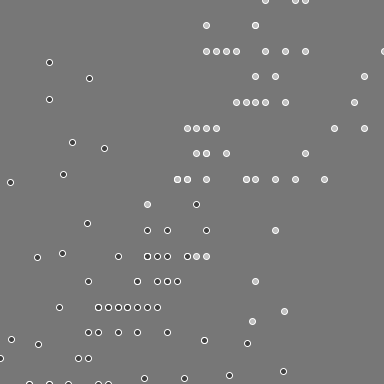}} \quad
        \fbox{\includegraphics[width=0.25\textwidth]{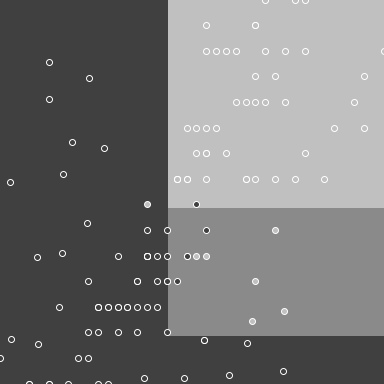}}
    	}
    \caption{\label{fig:motivation} Visualisation of predicted probabilities for the basic classifier described in Section~\ref{sec:artificial} (left) and the corresponding calibrated probabilities from a probability calibration tree (right) for a simple artificial dataset. The background colour indicates the probability estimates in the different regions.}
\end{figure}

To provide intuition on how a tree-based method can give improvements over existing methods for probability calibration, consider the case where we wish to calibrate probabilities produced by  a very basic classifier that always predicts the prior class distribution of the training data for every test instance. Platt scaling and isotonic regression are \textit{global} calibration methods,~\textit{i.e.}, they apply the same calibration model to the whole of the input space. As such, they are unable to improve upon the baseline provided by this basic classifier, which gives the same, constant prediction for every test instance. On the other hand, a probability calibration tree is able to build different probability calibration models for different regions of the input space. Fitting a probability calibration tree enables creation of local calibration models that compensate for the high bias of the basic classifier, which clearly is preferable to a constant predictor in almost every situation. In fact, the model produced by our algorithm is essentially the same as an ordinary decision tree in this case~(Figure~\ref{fig:motivation}). Of course, this is an extreme example, but there are many more complex models in machine learning that exhibit bias and regions in the input space that can benefit from a more specialised calibration model than the rest of the data. Probability calibration trees provide a way to identify these regions automatically and fit local calibration models to them. 

\section{\label{sec:experiments}Experiments}
In this section, we present results for experiments run on a range of UCI datasets. We compare the RMSE of probabilities calibrated with probability calibration trees to that by Platt scaling and isotonic regression for a number of base learners---naive Bayes, boosted stumps, boosted decision trees and SVMs. We do not compare to the other more recent methods mentioned in Section~\ref{sec:other_probability_calibration_methods}, which are methods of improving global calibration models. It would be interesting to apply these improvements to the local calibration models in probability calibration trees as an area of future work. We also present reliability diagrams~\citep{degroot1983comparison} for five datasets to qualitatively show the efficacy of our method. 

\subsection{Experiments on UCI Datasets}
\begin{table}[t]
\centering
\scriptsize
\caption{\label{tab:datasets} UCI datasets used in our experiments.}
\begin{tabular}{lccclccc}
\toprule
Dataset             & Instances & Attributes & Classes & Dataset & Instances & Attributes & Classes\\
\midrule
audiology       & 226   & 68  & 24 & news-popularity & 39644 & 59  & 2 \\
bankruptcy      & 10503 & 64  & 2  & nursery         & 12960 & 8   & 5 \\
colposcopy      & 287   & 62  & 2  & optdigits       & 5620  & 64  & 10\\
credit-rating   & 690   & 15  & 2  & page-blocks     & 5473  & 10  & 5 \\
cylinder-bands  & 512   & 39  & 2  & pendigits       & 10992 & 16  & 10\\
german-credit   & 1000  & 20  & 2  & phishing        & 1353  & 10  & 3 \\
hand-postures   & 78095 & 39  & 5  & pima-diabetes   & 768   & 8   & 2 \\
htru2           & 17898 & 8   & 2  & segment         & 2310  & 20  & 7 \\
kr-vs-kp        & 3196  & 36  & 2  & shuttle         & 58000 & 9   & 7 \\
led24           & 5000  & 24  & 10 & sick            & 3772  & 29  & 2 \\
mfeat-factors   & 2000  & 216 & 10 & spambase        & 4601  & 57  & 2 \\
mfeat-fourier   & 2000  & 76  & 10 & taiwan-credit   & 30000 & 23  & 2 \\
mfeat-karhunen  & 2000  & 64  & 10 & tic-tac-toe     & 958   & 9   & 2 \\
mfeat-morph     & 2000  & 6   & 10 & vote            & 435   & 16  & 2 \\
mfeat-pixel     & 2000  & 240 & 10 & vowel           & 990   & 14  & 10\\
mice-protein    & 1080  & 80  & 8  & yeast           & 1484  & 8   & 10\\
\bottomrule
\end{tabular}
\end{table}

We present results for 32 UCI datasets~\citep{Lichman2013}, listed in Table~\ref{tab:datasets}. These include 24 classic UCI datasets as well as eight more recently published datasets, which we briefly describe below.

\begin{description}
  \item[bankruptcy:] This dataset is about bankruptcy prediction of Polish companies~\citep{zikeba2016ensemble}. The classes are heavily unbalanced, and most features exhibit major outliers. 
  
  \item[colposcopy:] This dataset explores the subjective quality assessment of digital colposcopies. It contains features extracted from images of colposcopies~\citep{fernandes2017transfer}.
  
  \item[htru2:] This dataset describes a sample of pulsar candidates collected during the High Time Resolution Universe survey~\citep{lyon2016fifty}.
  
  \item[hand-postures:] Five types of hand postures from 12 users were recorded using unlabeled markers attached to fingers of a glove in a motion capture environment~\citep{gardner20143d}. Due to resolution and occlusion, missing values are common. 
  
  \item[mice-protein:] This dataset explores expression levels of 77 proteins measured in the cerebral cortex of mice exposed to context fear conditioning, a task used to assess associative learning~\citep{higuera2015self}.
  
  \item[news-popularity:] This dataset summarizes a heterogeneous set of features about articles published by Mashable in a period of two years~\citep{fernandes2015proactive}. The goal is to predict whether or not an article will be shared above a certain number of times.

  \item[phishing:] This dataset is about detecting websites that have been set up as phishing scams~\citep{abdelhamid2014phishing}.
  
  \item[taiwan-credit:] This dataset is about predicting if clients will default on their next payment, based on demographic data and their payment history~\citep{yeh2009comparisons}. 
\end{description}

\subsubsection{Experimental Setup}
We chose the four specific base learners as they are especially susceptible to producing poor probability estimates~\citep{niculescu2005predicting}, so they are more likely to benefit from a probability calibration scheme. When using isotonic regression on multiclass datasets, we use the one-vs-rest method to decompose the problem into several binary problems~\citep{zadrozny2002transforming, frank1998using}.

We used an existing implementation of isotonic regression, and implemented the probability calibration tree and Platt scaling in the WEKA framework~\citep{hall2009weka}. The algorithms we implemented are available from the WEKA package manager as \texttt{plattScaling} and \texttt{probabilityCalibrationTrees}. To make the comparison fair, we also transformed the input probability estimates to log-odds for Platt scaling.

The values in our results tables are the average of 10 runs, each run being the result of stratified 10-fold cross-validation. Note that the test sets in these cross-validation runs are not involved in the calibration process in any way. In our results tables, a filled circle ($\bullet$) indicates that our method provides a statistically significant improvement, and an open circle ($\circ$) indicates statistically significant degradation. A $p$-value of $0.01$ in conjunction with a corrected resampled $t$-test~\citep{nadeau2000inference} was used for all of our experiments to establish statistical significance. Note that we use a {\em corrected} version of the paired $t$-test that was shown to have Type I error at the significance level, and a conservative setting for the significance level (1\%). 

\subsubsection{Experimental Results}
\begin{table}[t]
\caption{\label{tab:naive_bayes}RMSE of each probability calibration method when calibrating probability estimates from naive Bayes.}
\footnotesize
{\centering \begin{tabular}{lcc@{\hspace{0.1cm}}cc@{\hspace{0.1cm}}cclcc@{\hspace{0.1cm}}cc@{\hspace{0.1cm}}cc@{\hspace{0.1cm}}cr@{\hspace{0.1cm}}c}
\toprule
Dataset 		& PCT  & PS     &           & IR    & 			& & Dataset 		& PCT   & PS    &           & IR    & 		 \\
\midrule
audiology       & 0.136 & 0.152 &           & 0.128 &           & & news-popularity & 0.471 & 0.491 & $\bullet$ & 0.483 & $\bullet$\\
bankruptcy      & 0.180 & 0.212 & $\bullet$ & 0.212 & $\bullet$ & & nursery         & 0.137 & 0.152 & $\bullet$ & 0.158 & $\bullet$\\
colposcopy      & 0.426 & 0.424 &           & 0.405 &           & & optdigits       & 0.123 & 0.170 & $\bullet$ & 0.116 & $\circ$  \\
credit-rating   & 0.322 & 0.369 & $\bullet$ & 0.354 & $\bullet$ & & page-blocks     & 0.104 & 0.152 & $\bullet$ & 0.128 & $\bullet$\\
cylinder-bands  & 0.400 & 0.420 & $\bullet$ & 0.416 &           & & pendigits       & 0.076 & 0.141 & $\bullet$ & 0.143 & $\bullet$\\
german-credit   & 0.410 & 0.413 &           & 0.410 &           & & phishing        & 0.240 & 0.273 & $\bullet$ & 0.274 & $\bullet$\\
hand-postures   & 0.120 & 0.312 & $\bullet$ & 0.259 & $\bullet$ & & pima-diabetes   & 0.410 & 0.412 &           & 0.410 &          \\
htru2           & 0.134 & 0.163 & $\bullet$ & 0.160 & $\bullet$ & & segment         & 0.112 & 0.213 & $\bullet$ & 0.166 & $\bullet$\\
kr-vs-kp        & 0.080 & 0.296 & $\bullet$ & 0.296 & $\bullet$ & & shuttle         & 0.020 & 0.108 & $\bullet$ & 0.094 & $\bullet$\\
led24           & 0.194 & 0.194 &           & 0.194 & $\circ$   & & sick            & 0.100 & 0.184 & $\bullet$ & 0.178 & $\bullet$\\
mfeat-factors   & 0.113 & 0.127 & $\bullet$ & 0.106 &           & & spambase        & 0.242 & 0.337 & $\bullet$ & 0.286 & $\bullet$\\
mfeat-fourier   & 0.164 & 0.174 & $\bullet$ & 0.173 & $\bullet$ & & taiwan-credit   & 0.369 & 0.380 & $\bullet$ & 0.375 & $\bullet$\\
mfeat-karhunen  & 0.084 & 0.094 & $\bullet$ & 0.097 & $\bullet$ & & tic-tac-toe     & 0.359 & 0.431 & $\bullet$ & 0.413 & $\bullet$\\
mfeat-morph     & 0.195 & 0.224 & $\bullet$ & 0.196 &           & & vote            & 0.189 & 0.254 & $\bullet$ & 0.250 & $\bullet$\\
mfeat-pixel     & 0.081 & 0.086 & $\bullet$ & 0.100 & $\bullet$ & & vowel           & 0.151 & 0.207 & $\bullet$ & 0.208 & $\bullet$\\
mice-protein    & 0.015 & 0.209 & $\bullet$ & 0.140 & $\bullet$ & & yeast           & 0.239 & 0.250 & $\bullet$ & 0.236 & $\circ$  \\
\bottomrule
\multicolumn{12}{c}{$\bullet$, $\circ$ statistically significant improvement or degradation, $p = 0.01$}\\
\end{tabular} \footnotesize \par}
\end{table}

\begin{table}[t]
\caption{\label{tab:boosted_stumps}RMSE of each probability calibration method when calibrating probability estimates from boosted stumps.}
\footnotesize
{\centering \begin{tabular}{lcc@{\hspace{0.1cm}}cc@{\hspace{0.1cm}}cclcc@{\hspace{0.1cm}}cc@{\hspace{0.1cm}}cc@{\hspace{0.1cm}}cr@{\hspace{0.1cm}}c}

\toprule
Dataset 			& PCT  & PS  &         & IR  & 			  & & Dataset 			& PCT  & PS  &         & IR  & 		 \\
\midrule
audiology       & 0.131 & 0.122 &           & 0.121 &           & & news-popularity & 0.461 & 0.461 &           & 0.461 &          \\
bankruptcy      & 0.174 & 0.186 & $\bullet$ & 0.186 & $\bullet$ & & nursery         & 0.107 & 0.146 & $\bullet$ & 0.142 & $\bullet$\\
colposcopy      & 0.410 & 0.406 &           & 0.407 &           & & optdigits       & 0.064 & 0.064 &           & 0.065 &          \\
credit-rating   & 0.318 & 0.329 &           & 0.328 &           & & page-blocks     & 0.095 & 0.095 &           & 0.095 &          \\
cylinder-bands  & 0.385 & 0.384 &           & 0.387 &           & & pendigits       & 0.053 & 0.057 & $\bullet$ & 0.056 & $\bullet$\\
german-credit   & 0.411 & 0.412 &           & 0.412 &           & & phishing        & 0.235 & 0.257 & $\bullet$ & 0.256 & $\bullet$\\
hand-postures   & 0.074 & 0.090 & $\bullet$ & 0.089 & $\bullet$ & & pima-diabetes   & 0.404 & 0.405 &           & 0.406 &          \\
htru2           & 0.131 & 0.131 &           & 0.131 &           & & segment         & 0.122 & 0.143 &           & 0.141 &          \\
kr-vs-kp        & 0.085 & 0.157 & $\bullet$ & 0.153 & $\bullet$ & & shuttle         & 0.005 & 0.008 & $\bullet$ & 0.005 &          \\
led24           & 0.195 & 0.195 &           & 0.195 &           & & sick            & 0.102 & 0.133 & $\bullet$ & 0.132 & $\bullet$\\
mfeat-factors   & 0.068 & 0.066 &           & 0.066 &           & & spambase        & 0.202 & 0.203 &           & 0.203 &          \\
mfeat-fourier   & 0.156 & 0.156 &           & 0.159 &           & & taiwan-credit   & 0.368 & 0.369 & $\bullet$ & 0.368 & $\bullet$\\
mfeat-karhunen  & 0.083 & 0.081 &           & 0.082 &           & & tic-tac-toe     & 0.143 & 0.176 & $\bullet$ & 0.169 &          \\
mfeat-morph     & 0.181 & 0.181 &           & 0.181 &           & & vote            & 0.178 & 0.182 &           & 0.184 &          \\
mfeat-pixel     & 0.080 & 0.079 &           & 0.080 &           & & vowel           & 0.115 & 0.119 &           & 0.125 &          \\
mice-protein    & 0.000 & 0.008 & $\bullet$ & 0.004 &           & & yeast           & 0.234 & 0.234 &           & 0.235 &          \\
\bottomrule
\multicolumn{12}{c}{$\bullet$, $\circ$ statistically significant improvement or degradation, $p=0.01$}\\
\end{tabular} \footnotesize \par}
\end{table}

Table~\ref{tab:naive_bayes} shows the results of performing calibration on the probability estimates produced by naive Bayes. Naive Bayes is known to produce particularly poor probability estimates as it makes unrealistic independence assumptions about the attributes. We used the default settings in WEKA for hyperparameters. It is clear from the results table that probability calibration trees typically outperform both Platt scaling and isotonic regression when calibrating probabilities produced by naive Bayes. In every case, our method either equals or achieves a statistically significantly lower error than Platt scaling. Isotonic regression is superior to our method on three datasets, but is outperformed in many others. We can see that probability calibration trees work particularly well compared with Platt scaling and isotonic regression when naive Bayes is used as the base learner. This is because naive Bayes is a less powerful model than the other base learners we tested, so the calibration process can benefit more from the additional tree structure.

Tables~\ref{tab:boosted_stumps} and~\ref{tab:boosted_trees} show the results of performing calibration on the probability estimates produced by boosted decision stumps and boosted decision trees, respectively. The calibration curves of boosted decision stumps and trees typically exhibit a sigmoid shape~\citep{niculescu2005predicting}, so we would expect probability calibration trees and Platt scaling to work well on these estimates. We used the LogitBoost algorithm to boost 100 stumps and 100 REPTrees from WEKA. We set the maximum depth of the decision trees to three, and the minimum number of instances at the leaf nodes to zero. We also disabled automatic pruning for the trees used as the base learners. As with naive Bayes, our method either performs as well as, or significantly better than, Platt scaling on every dataset we tested on. However, for these experiments, our method also either outperforms or equals the performance of isotonic regression on every dataset we tested. Even though the numbers of wins and losses are not as dramatic as for naive Bayes, probability calibration trees still surpass the performance of the other methods on average.

\begin{table}[t]
\caption{\label{tab:boosted_trees}RMSE of each probability calibration method when calibrating probability estimates from boosted trees.}
\footnotesize
{\centering \begin{tabular}{lcc@{\hspace{0.1cm}}cc@{\hspace{0.1cm}}cclcc@{\hspace{0.1cm}}cc@{\hspace{0.1cm}}cc@{\hspace{0.1cm}}cr@{\hspace{0.1cm}}c}

\toprule
Dataset 			& PCT  & PS  &         & IR  & 			  & & Dataset 			& PCT  & PS  &         & IR  & 		 \\
\midrule
audiology       & 0.133 & 0.131 &           & 0.124 &           & & news-popularity & 0.465 & 0.469 & $\bullet$ & 0.468 & $\bullet$\\
bankruptcy      & 0.168 & 0.182 & $\bullet$ & 0.182 & $\bullet$ & & nursery         & 0.005 & 0.018 & $\bullet$ & 0.006 &          \\
colposcopy      & 0.402 & 0.401 &           & 0.404 &           & & optdigits       & 0.046 & 0.045 &           & 0.045 &          \\
credit-rating   & 0.319 & 0.349 & $\bullet$ & 0.352 & $\bullet$ & & page-blocks     & 0.094 & 0.094 &           & 0.095 &          \\
cylinder-bands  & 0.449 & 0.482 & $\bullet$ & 0.431 &           & & pendigits       & 0.029 & 0.028 &           & 0.029 &          \\
german-credit   & 0.413 & 0.421 &           & 0.423 &           & & phishing        & 0.228 & 0.239 &           & 0.239 &          \\
hand-postures   & 0.045 & 0.056 & $\bullet$ & 0.051 & $\bullet$ & & pima-diabetes   & 0.410 & 0.422 &           & 0.424 &          \\
htru2           & 0.131 & 0.137 & $\bullet$ & 0.137 & $\bullet$ & & segment         & 0.055 & 0.053 &           & 0.055 &          \\
kr-vs-kp        & 0.043 & 0.043 &           & 0.041 &           & & shuttle         & 0.004 & 0.005 &           & 0.003 &          \\
led24           & 0.208 & 0.213 & $\bullet$ & 0.218 & $\bullet$ & & sick            & 0.086 & 0.083 &           & 0.085 &          \\
mfeat-factors   & 0.062 & 0.060 &           & 0.060 &           & & spambase        & 0.192 & 0.192 &           & 0.192 &          \\
mfeat-fourier   & 0.145 & 0.144 &           & 0.148 &           & & taiwan-credit   & 0.369 & 0.380 & $\bullet$ & 0.378 & $\bullet$\\
mfeat-karhunen  & 0.073 & 0.071 &           & 0.072 &           & & tic-tac-toe     & 0.025 & 0.036 &           & 0.025 &          \\
mfeat-morph     & 0.178 & 0.177 &           & 0.179 &           & & vote            & 0.191 & 0.204 &           & 0.202 &          \\
mfeat-pixel     & 0.080 & 0.078 &           & 0.079 &           & & vowel           & 0.093 & 0.092 &           & 0.096 &          \\
mice-protein    & 0.008 & 0.012 &           & 0.002 &           & & yeast           & 0.240 & 0.239 &           & 0.241 &          \\
\bottomrule
\multicolumn{12}{c}{$\bullet$, $\circ$ statistically significant improvement or degradation, $p=0.01$}\\
\end{tabular} \footnotesize \par}
\end{table}

Table~\ref{tab:svm} shows the results of performing calibration on the outputs produced by SVMs with RBF kernels. We performed a 2-fold grid search over the $C$ and $\gamma$ hyperparameters, ranging from $10^{-2}$ to $10^2$ (in increments of powers of 10) for each value, to optimise the accuracy of the SVM before calibrating. For the multiclass datasets with 5000 instances or less, the grid search was performed on a random sample of 20\% of the training data. For those multiclass datasets with more than 5000 instances, a 10\% random sample was taken for the grid search. Note that the grid search is performed on each fold independently, and the test data is not included in the hyperparameter optimisation process. We used the one-vs-rest technique to apply the SVMs to the multiclass datasets. Note that the output of an SVM is not a probability estimate, and as such, the vector of values passed to the calibration methods is not a probability distribution like for the other base learners. In order to produce a vector of values to input into the calibration model, we took the outputs of each one-vs-rest model and concatenated them together:

\begin{equation}
	\textrm{vector} = \big[ S_1(x), S_2(x), \dots, S_m(x) \big]
\end{equation}

where $S_i(x)$ is the output of the SVM trained to differentiate class $i$ from the rest of the classes. Furthermore, we did not apply the log-odds transformation (Equation~\ref{eqn:logodds}) to these values for any calibration method. Calibration results for SVMs in Table~\ref{tab:svm} show that again, our method performs better on average than Platt scaling and isotonic regression, with several wins and no losses for each method.

\begin{table}[t]
\caption{\label{tab:svm}RMSE of each probability calibration method when calibrating outputs from SVMs with RBF kernels.}
\footnotesize
{\centering \begin{tabular}
{lcc@{\hspace{0.1cm}}cc@{\hspace{0.1cm}}cclcc@{\hspace{0.1cm}}cc@{\hspace{0.1cm}}cc@{\hspace{0.1cm}}cr@{\hspace{0.1cm}}c}

\toprule
Dataset 			& PCT   & PS    &           & IR    & 			& & Dataset 			  & PCT   & PS    &           & IR    & 		 \\
\midrule
audiology       & 0.123 & 0.126 &           & 0.116 &           & & news-popularity & 0.468 & 0.474 & $\bullet$ & 0.470 & $\bullet$\\
bankruptcy      & 0.180 & 0.212 & $\bullet$ & 0.207 & $\bullet$ & & nursery         & 0.009 & 0.066 & $\bullet$ & 0.034 &          \\
colposcopy      & 0.400 & 0.396 &           & 0.397 &           & & optdigits       & 0.037 & 0.038 &           & 0.038 &          \\
credit-rating   & 0.328 & 0.327 &           & 0.332 &           & & page-blocks     & 0.121 & 0.118 &           & 0.102 &          \\
cylinder-bands  & 0.338 & 0.344 &           & 0.349 &           & & pendigits       & 0.027 & 0.027 &           & 0.026 &          \\
german-credit   & 0.407 & 0.406 &           & 0.408 &           & & phishing        & 0.244 & 0.244 &           & 0.241 &          \\
hand-postures   & 0.062 & 0.076 & $\bullet$ & 0.063 &           & & pima-diabetes   & 0.396 & 0.394 &           & 0.396 &          \\
htru2           & 0.129 & 0.133 & $\bullet$ & 0.129 &           & & segment         & 0.085 & 0.086 &           & 0.086 &          \\
kr-vs-kp        & 0.048 & 0.050 &           & 0.049 &           & & shuttle         & 0.030 & 0.083 & $\bullet$ & 0.020 &          \\
led24           & 0.197 & 0.198 &           & 0.199 &           & & sick            & 0.102 & 0.167 & $\bullet$ & 0.162 & $\bullet$\\
mfeat-factors   & 0.055 & 0.059 &           & 0.056 &           & & spambase        & 0.223 & 0.235 & $\bullet$ & 0.228 &          \\
mfeat-fourier   & 0.147 & 0.148 &           & 0.155 & $\bullet$ & & taiwan-credit   & 0.371 & 0.379 & $\bullet$ & 0.374 & $\bullet$\\
mfeat-karhunen  & 0.069 & 0.072 &           & 0.072 &           & & tic-tac-toe     & 0.162 & 0.105 &           & 0.142 &          \\
mfeat-morph     & 0.187 & 0.182 &           & 0.184 &           & & vote            & 0.177 & 0.174 &           & 0.178 &          \\
mfeat-pixel     & 0.059 & 0.055 &           & 0.061 &           & & vowel           & 0.054 & 0.075 &           & 0.066 &          \\
mice-protein    & 0.000 & 0.010 &           & 0.000 &           & & yeast           & 0.239 & 0.239 &           & 0.235 &          \\
\hline
\multicolumn{12}{c}{$\bullet$, $\circ$ statistically significant improvement or degradation}\\
\end{tabular} \footnotesize \par}
\end{table}

Finally, we summarise these results as a series of two-tailed sign tests in Table~\ref{tab:signtest}. Every test except one is significant at $p < 0.01$, with many of the $p$-values being much lower. Probability calibration trees have zero losses on nearly every comparison. It can be seen that our method has a fairly large number of draws. Draws with Platt scaling tend to have almost identical RMSE---this indicates that for these datasets, the probability calibration tree was likely pruned back to the root node, and so a global Platt scaling model is appropriate for calibrating the output scores from the corresponding base learner.  

\begin{table}[h]
\centering
\caption{\label{tab:signtest}Sign test comparing statistically significant wins (W), draws (D) and losses (L) for probability calibration trees against Platt scaling and isotonic regression for each base model from our experiments.}
\label{my-label}
\begin{tabular}{rccccccccc}
 \addlinespace[-\aboverulesep] \cmidrule[\heavyrulewidth]{2-10}
               & \multicolumn{4}{c}{Platt Scaling} & & \multicolumn{4}{c}{Isotonic Regression} \\ 
               \cmidrule{2-5} \cmidrule{7-10} 
               & W   & D   & L  & $p$-value         & & W       & D       & L     & $p$-value      \\ \midrule
Naive Bayes    & 27  & 5   & 0  & \textless0.00001  & & 23      & 6       & 3     & 0.000088     \\
Boosted Stumps & 11  & 21  & 0  & 0.00091           & & 8       & 24      & 0     & 0.004678     \\
Boosted Trees  & 9   & 23  & 0  & 0.0027            & & 7       & 25      & 0     & 0.008151     \\
SVM            & 9   & 23  & 0  & 0.0027            & & 5       & 27      & 0     & 0.025347     \\ \bottomrule
\end{tabular}
\end{table}

\subsection{Reliability Diagrams}

Reliability diagrams~\citep{degroot1983comparison} are plots that compare the estimated probabilities produced by a (binary) classifier to their empirical distribution, and are a way to visualise the performance of probability calibration methods. Initially, the output space is discretised into a number of bins. The test instances are then grouped into these bins according to their associated predicted probabilities from the classifier. Finally, the average predicted probability for each bin is plotted against the true percentage of positive examples in the bin. Well-calibrated probabilities should result in the data points falling near the diagonal line.

After analysing the distribution of calibrated probabilities obtained by applying calibration methods to naive Bayes for a selection of the datasets, we observed that many of the bins in the center of the plot had very few instances if the bins were chosen with equal width, so we instead chose bins with equal frequency. A maximum of 30 bins was used, although most plots have fewer points due to ties.

	\begin{figure}[p!]
	    \centering
	    \includegraphics[width=0.95\textwidth]{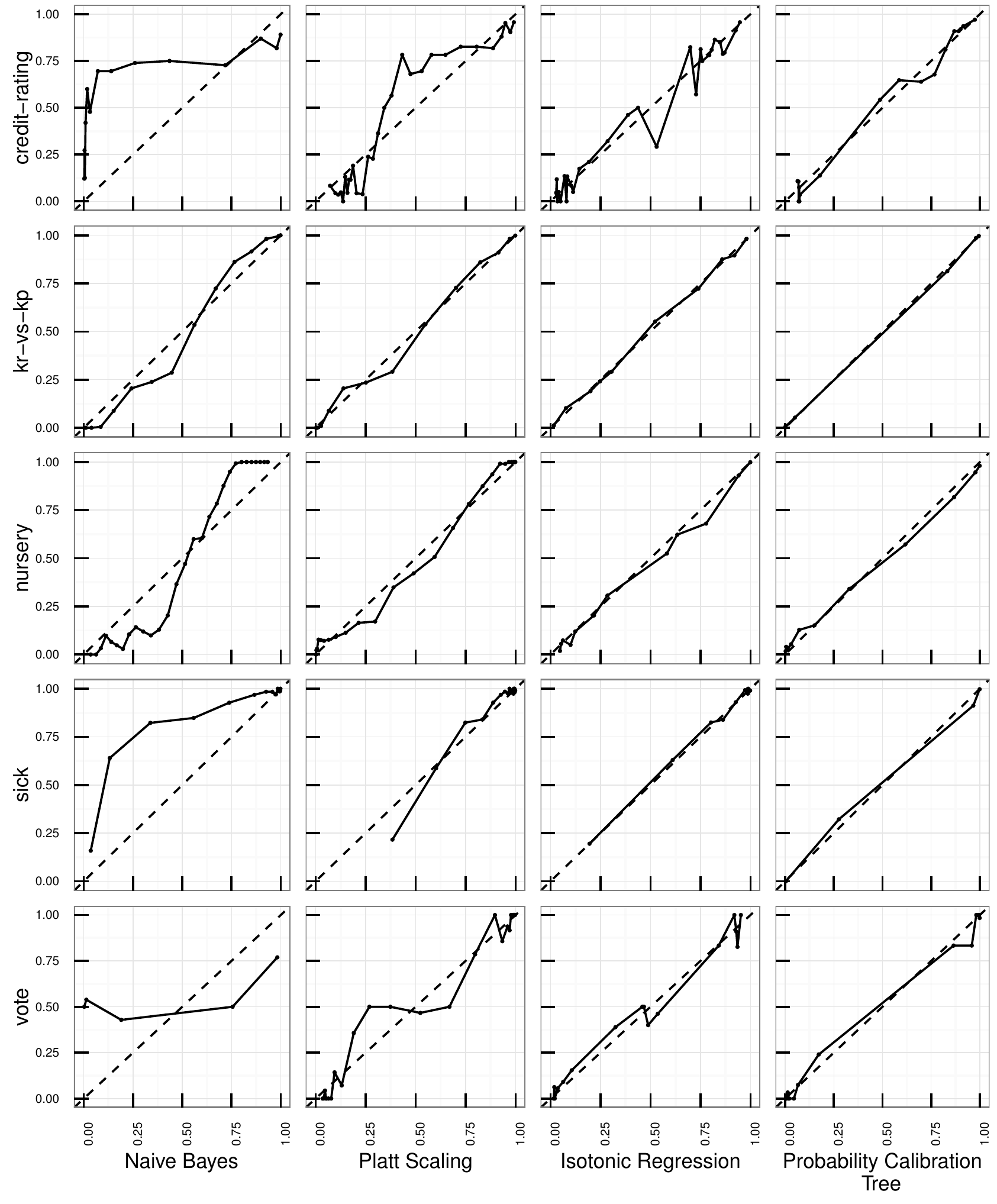}~~~
	    \caption{\label{fig:reliability_diagrams}Reliability diagrams for each calibration method operating on probabilities produced by naive Bayes. Each row shows plots for a single dataset, while each column shows the plots for a single method. The first column shows the original probabilities estimated by naive Bayes. The $x$-axis represents the predicted probabilities, while the $y$-axis represents the empirical probabilities of each bin.}
	\end{figure}

Figure~\ref{fig:reliability_diagrams} shows reliability diagrams for the credit-rating, kr-vs-kp, nursery (only priority and spec\_prior classes), sick and vote datasets for each of the calibration methods, as well as naive Bayes. We can see that the original probabilities from naive Bayes are very poorly calibrated for all five datasets. Note that the sick dataset is heavily imbalanced in favour of the positive class, so the mean of the first equal-frequency bin is a relatively high value compared to the other datasets. 

It can be seen that Platt scaling typically exhibits some of the features of the reliability curve of the original classifier. Even though its curve is much closer to the diagonal, the limitations of fitting a global sigmoid model to noisy estimates cause the general shape to remain similar in appearance. Isotonic regression does not have this problem, but the reliability curve appears quite jagged and crosses over the diagonal line many times. This is due to the piecewise constant nature of the calibration function, which results in many probability estimates being calibrated to the same value. Finally, the reliability curves of probability calibration trees generally appear smooth and follow the diagonal line closely, demonstrating that probability calibration trees are able to calibrate probabilities well in comparison.

\section{\label{sec:conclusion}Conclusion}
We have presented a method for probability calibration---induction of probability calibration trees---that is derived from the process of growing logistic model trees. The original predictor attributes are used for splitting the data to grow the tree, while the base learners' output scores are used to fit logistic regression models to the nodes of the tree. In this manner, probability calibration trees are able to split the input space into regions where different calibration models can be trained locally to improve overall calibration performance. Our method has been shown to substantially outperform Platt scaling and isotonic regression when applied to probabilities from naive Bayes, and to perform better on average for calibrating boosted decision trees, boosted stumps and SVMs. 

\acks{This research was supported by the Marsden Fund Council from Government funding, administered by the Royal Society of New Zealand.}

\bibliography{leathart17}
\end{document}